\documentclass[sigconf]{acmart}

\usepackage{booktabs} 
\usepackage{subfigure}
\usepackage{multirow}

\copyrightyear{2019} 
\acmYear{2019} 
\setcopyright{acmcopyright}
\acmConference[KDD '19] {The 25th ACM SIGKDD Conference on Knowledge Discovery and Data Mining}{August 4--8, 2019}{Anchorage, AK, USA}
\acmBooktitle{The 25th ACM SIGKDD Conference on Knowledge Discovery and Data Mining (KDD '19), August 4--8, 2019, Anchorage, AK, USA}
\acmPrice{15.00}
\acmDOI{10.1145/XXXXXX.XXXXXX}
\acmISBN{978-1-4503-6201-6/19/08} 

\begin{document}
\title{FoodAI: Food Image Recognition via Deep Learning\\ for Smart Food Logging}

\author{Doyen Sahoo$^{1}$, Wang Hao$^{1}$, Shu Ke$^{1}$, Wu Xiongwei$^{1}$, Hung Le$^{1}$, \\Palakorn Achananuparp$^{1}$, Ee-Peng Lim$^{1}$, Steven C. H. Hoi$^{1,2}$}
\affiliation{
  \institution{$^{1}$Living Analytics Research Centre (LARC), School of Information Systems,Singapore Management University}
  \institution{$^{2}$Salesforce Research Asia}
}
\email{{doyens,hwang,keshu,xwwu.2015,hungle.2018,palakorna,eplim,chhoi}@smu.edu.sg}

\begin{abstract}
	An important aspect of health monitoring is effective logging of food consumption. This can help management of diet-related diseases like obesity, diabetes, and even cardiovascular diseases. Moreover, food logging can help fitness enthusiasts, and people who wanting to achieve a target weight. However, food-logging is cumbersome, and requires not only taking additional effort to note down the food item consumed regularly, but also sufficient knowledge of the food item consumed (which is difficult due to the availability of a wide variety of cuisines). With increasing reliance on smart devices, we exploit the convenience offered through the use of smart phones and propose a smart-food logging system: FoodAI\footnote{www.foodai.org}, which offers state-of-the-art deep-learning based image recognition capabilities. FoodAI has been developed in Singapore and is particularly focused on food items commonly consumed in Singapore. FoodAI models were trained on a corpus of 400,000 food images from 756 different classes. In this paper we present extensive analysis and insights into the development of this system. FoodAI has been deployed as an API service and is one of the components powering Healthy 365, a mobile app developed by Singapore's Heath Promotion Board. We have over 100 registered organizations (universities, companies, start-ups) subscribing to this service and actively receive several API requests a day. FoodAI has made food logging convenient, aiding smart consumption and a healthy lifestyle.
\end{abstract}

%

\ccsdesc[500]{Computing methodologies~Object recognition}
\ccsdesc[500]{Applied computing~Consumer health}

\keywords{Food Computing, Image Recognition, Smart Food Logging}

\maketitle

\section{Introduction}

Food consumption has a multifaceted impact on us, including health, culture, behavior, preferences and many other aspects \cite{min2018survey}. Particularly, food habits are among the main reasons for several health-related ailments prevalent in society. Improper consumption patterns can often lead to people becoming overweight or obese, which is further linked to chronic illnesses like diabetes and cardiovascular diseases. Tracking food consumption behavior is thus a critical requirement to not only help individuals to prevent diseases, but also for those suffering from these disease to manage their health better. There is an extensive prevalence of obesity and diabetes globally\footnote{www.who.int/news-room/fact-sheets/detail/diabetes}, and  there is a need to formulate strategies to counter these issues. In Singapore, where the FoodAI project has been undertaken, diabetes has been identified as major problem that needs addressing, and several steps have been taken towards this goal\footnote{www.diabetes.org.sg/}. There are many ways to manage diabetes (e.g. regular health check-ups, diligently following treatments, etc.)\footnote{www.niddk.nih.gov/health-information/diabetes/overview/managing-diabetes}, where a major component is just effective monitoring of diets. Moreover, effective monitoring of diets can serve as a good prevention against diet-related ailments, and also help fitness enthusiasts achieve their weight goals. 

A traditional approach to monitoring diets is maintaining a food journal, where we note down the food items everytime we consume them. Such approaches are tedious and require substantial manual effort from the users. While motivated individuals can sustain this habit for a while, many might give it up soon due to the inconvenience. Such an effort is also very inefficient, not only from the perspective of logging, but also from the perspective of analyzing the data. As a result, obtaining actionable insights in a data-driven manner becomes difficult. Another major concern with such a mechanism for logging is the assumption that the individual has sufficient knowledge about the food items they are consuming. Singapore is known for its highly diverse cuisines. Not only are there several local items, but also cuisines from all over the world are found here. This environment provides a wide array of novel food choices for consumers, but makes effective food logging difficult, as now the users also need to make an effort to identify the details of the unfamiliar food items they are consuming, including their nutritional content. This is particularly harder when the item names are in different languages or not very descriptive. 

\begin{figure}[h!]
	\centering
	\includegraphics[height=8cm,width=0.5\textwidth]{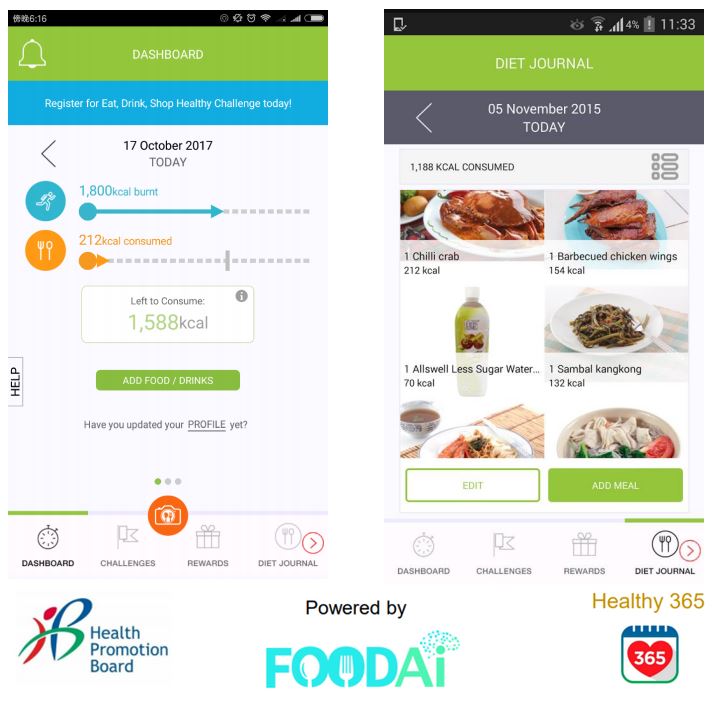}
	\caption{An example of food logging in a diet journal on the Healthy 365 App, which is developed and mantained by Health Promotion Board, and is powered by FoodAI.}
	\label{fig:log_example}
\end{figure}

With the growing reliance on smart and self-tracking devices, there has been an increasing demand for innovative technologies to ease the tracking behavior of individuals \cite{solanas2014smart}. One of the most prevalent examples is the usage of smart-phones and fitness bands to monitor physical activity (number of steps, distance traveled, exercise, etc.) \cite{el2015currently}. Following this trend, we propose to exploit the convenience of smart-phones, and empower them with FoodAI, our deep-learning based food-image recognition system, to build a smart food logging system. This approach overcomes the limitations of traditional logging techniques and allows for a very efficient and effective logging. The entire system architecture is deployed as an API service, with either a mobile app or a web-interface at the front-end. At the backend, we store the trained deep neural network model in the inference engine. There is also a database which stores the images captured by the users of the service, and the public food images collected by us during the web data crawling. During the offline phase, we collected and annotated a large scale dataset and trained a prediction model on this dataset using state-of-the-art image recognition deep learning models. Users access our food image recognition service through the API. One of our main partners is the Health Promotion Board of Singapore, who have developed Healthy365 App to help users maintain a healthy lifestyle. A major component of this app is a diet journal for users to track their food consumption. This diet journal is powered by FoodAI at the backend. When having a meal, the users can take an image of the meal, and log it down. A sample screenshot of how this app works is shown in Figure \ref{fig:log_example}. Over time, the users are able to monitor their eating habits, their caloric intake and outtake, etc. In the era of building a smart nation, FoodAI offers a solution to addressing smart consumption and healthy lifestyle for the society. 

FoodAI can recognize 756 different classes of foods. These items include main courses, drinks, as well as snacks. A food-image dataset of almost 400,000 images was crawled from public web search results and manually annotated for the purpose of building our training corpus. 100 classes from the 756 were collected with a specific focus on local food items commonly consumed in Singapore (>500 images per class). Extensive efforts were made to curate this dataset, and we also developed approaches to efficiently introduce new food classes into the system (or remove or merge existing classes). We used models pre-trained on ImageNet and fine tuned on our dataset. Based on our train-validation split, our approaches achieved over 83\% top-1 accuracy and 95\% top-5 accuracy. This model was then deployed as an API service, regularly receiving several API requests daily from more than 100 local and international organizations, universities, and start-ups. We frequently monitor these requests and study the performance of the model. We have conducted extensive qualitative and quantitative analysis of the usage patterns, and accordingly we keep making efforts to come up with actionable insights to continually improve the system.

\section{Related Work}

	Food computing \cite{min2018survey} has evolved as a popular research topic in recent years. Thanks to the proliferation of smart phones and social media, a large amount of food-related information (such as food images, cooking recipes, and food consumption logs) is often shared online. This has led to us having access to rich heterogeneous information for several important tasks. In particular, this gives the community an opportunity to conduct extensive analysis of food-related topics. Effectively utilizing food computing impacts our lives in multi-faceted ways, including our behavior, culture \cite{sajadmanesh2017kissing}, and health \cite{abbar2015you,mejova2015foodporn,ofli2017saki,achananuparp2018does}. In the past, such analysis has led to impact in medicine\cite{farinella2016retrieval}, biology\cite{batt2007food}, gastronomy \cite{ahn2011flavor,teng2012recipe}, agronomy \cite{hernandez2017search}, etc.

	One of the most important tasks is food image recognition using deep learning. Many of these techniques rely on the recent success of deep learning for visual recognition \cite{krizhevsky2012imagenet,lecun2015deep,goodfellow2016deep,he2016deep}, and use these state-of-the-art models to train a deep convolution network that can recognize a variety of food items. Using their feature extraction ability \cite{yosinski2014transferable}, researchers adapt pre-trained ImageNet \cite{deng2009imagenet} models to their own food datasets. Some of the recent examples include \cite{meyers2015im2calories,beijbom2015menu,liu2016deepfood,ming2018food,liu2018new}. Among various efforts to build food image recognition models \cite{min2018survey}, FoodAI has been trained on the largest food dataset for recognition tasks, with almost 400,000 images. 
	
	In addition to FoodAI, there are several existing commercial and academic food image recognition systems which can be employed to reduce the burdens of traditional mobile food journaling. These include CalorieMama\footnote{https://www.caloriemama.ai/}, AVA\footnote{https://eatwithava.com/}, 
	Salesforce Research's Food Image Model\footnote{https://metamind.readme.io/docs/food-image-model}, Google Vision API\footnote{https://cloud.google.com/vision/}, Amazon Rekognition\footnote{https://aws.amazon.com/rekognition/}, and many others. To the best of our knowledge, FoodAI is the most comprehensive food image recognition solution, with an ability to recognize over 756 different visual food categories (over 1,166 food items), specifically covering a wide variety of Southeast Asian cuisines and Asian cuisines in general.
	
	

\section{Food-AI}

	We now present our proposed FoodAI. We first describe data collection challenges, model training and addressing class imbalance issues. Then, we present extensive experiments and qualitative analysis to shed light onto the insights for deploying such a technology in the real world. We obtain insights from the development environment (analysis on our data collected for training), and the production environment (analysis of query data from users).
	
	\subsection{Constructing Food Image Dataset}
	
	During the early days, the primary objective was to develop a robust dataset of several food images. Special attention was given to food items commonly consumed in Singapore. 
	We first defined 152 "super categories" representing generic types of foods and drinks. Some of these super categories are: \textit{Beer, Fried Rice, Grilled Chicken, Ice Cream, etc.}. For each of these super categories, we identified several \textit{food-items} that likely belong to the category. For example, the \textit{Fruit} super category would have pineapple, jackfruit, and lychee as the food items. In this manner, we identified a total of 1,166 different food items. Ideally, this is the total number of classes. However, it turns out many items are not visually distinguishable (e.g., coffee with sugar and coffee without sugar are visually indistinguishable, but the difference is important as it significantly affects the users' total caloric intake). Thus, we introduced a notion of \textit{Visual Foods} as a way to further group the items in a specific category according to their visual similarity. By visually inspecting the images and consulting with domain experts, we merged the 1,166 different food items to 756 visual food categories. We also added a category for non-food items, for which we randomly sampled about 10,000 different images from ImageNet dataset \cite{deng2009imagenet}. The prediction model would be trained to make predictions on these visual food categories (or classes). In the case where finer grained predictions were needed at the application level (e.g., during food logging), the users would have the option to manually select a sub-item from the prediction made. For example, given an image of coffee beverage, the model would predict \textit{Coffee With Milk} and the user can further choose whether it was \textit{With Sugar} or \textit{Without Sugar}. As our target market was primarily Singapore, we laid special focus on foods commonly consumed in Singapore. Out of the 152 categories and 756 visual food categories, 8 categories (e.g. \textit{Indian, Chinese, Desserts, Malay, etc.}) and 100 visual foods are tailored to this purpose. During the dataset collection, we ensured that we had at least 500 images for each of these 100 visual foods (totaling to 65,855 images).
	
\begin{figure*}[h]
		\centering
		\includegraphics[height=4cm,width=0.8\textwidth]{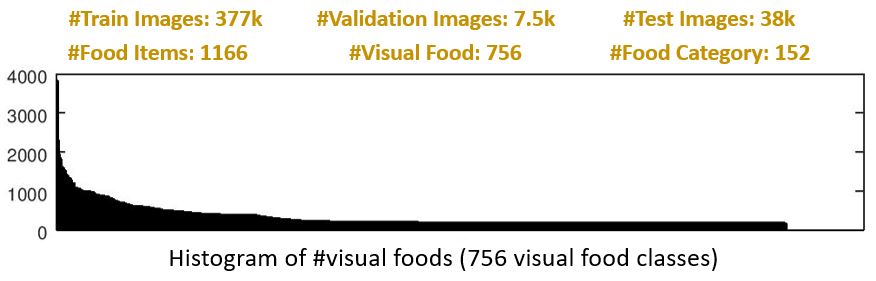}
		\caption{Details on the FoodAI-756 dataset. We have a total of about 400k images, and 756 classes (or visual food categories). Some of these classes encompass multiple food items (e.g. coffee would encompass both coffee with and without sugar.)}
		\label{fig:dataDistribution}
	\end{figure*}

	The images were collected by crawling from Google, Bing, Instagram, Flickr, Facebook and other social media, for each of the food categories. These were manually vetted to confirm whether each image crawled indeed belonged to the food class being searched for. Based on requests from the stakeholders and our continued efforts to improve FoodAI, we continue to update the dataset by including new images and visual food classes. At present, we have 756 visual foods, comprising about 400,000 images in total. We have a minimum of 174 images and maximum of 2,312 images per visual food. We then split the dataset into train, validation, and test data, for training and evaluating the model. The dataset (henceforth \textit{FoodAI-756}) is summarized in Figure \ref{fig:dataDistribution}.

	\subsubsection{Food Annotation Management System (FAMS)}
	
	\begin{figure*}[h]
		\centering
		\includegraphics[height=6cm,width=0.75\textwidth]{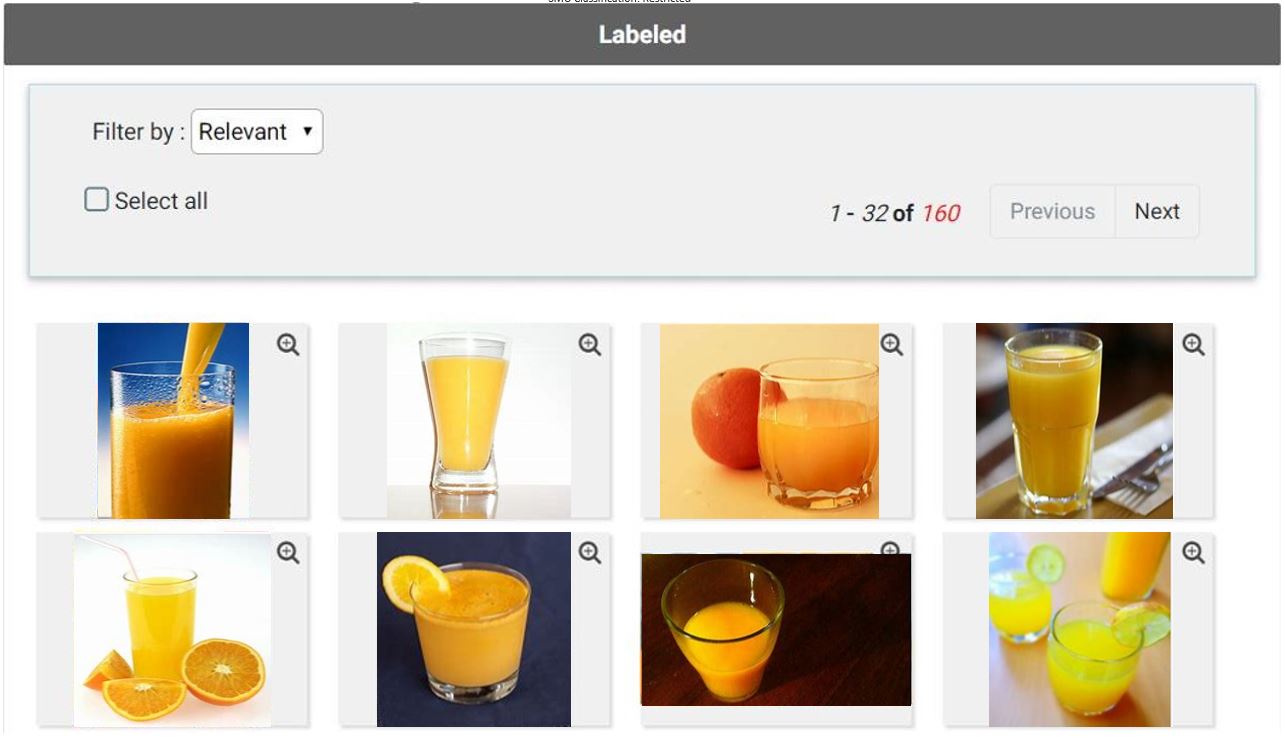}
		\caption{Food Annotation Management System (FAMS). An efficient crawler developed to facilitate effective collection of new data and new food classes based on the requirements from various stakeholders. Here we show an example of orange juice.}
		\label{fig:fams}
	\end{figure*}

	The original image crawling process was labor-intensive and inefficient, requiring users to manually query a search item and select appropriate images to download a given food category. To alleviate these and streamline the process, we developed Food Annotation Management System (FAMS), a web-based tool for automatic crawling and annotating food images. An annotator will define and submit a list of keywords (i.e., food items) through FAMS. The back-end crawler will then retrieve several thumbnail images and present them to the annotator in FAMS. The number of images retrieved is based on the input provided by the user. The annotator is then able to view several thumbnails and is able to quickly identify relevant images for the given keyword (e.g., by checking all, and unchecking the few irrelevant images). After this, the annotator can confirm the labeling and the full-size images are automatically downloaded and saved to the database. An example of collecting images for \textit{Orange Juice} is shown in Figure \ref{fig:fams}. There are two important roles in FAMS, manager and annotator. The manager manages the annotation processes and assigns annotation tasks to one or more annotators. Annotators complete the tasks assigned by the manager. After the manager has confirmed the annotation results, FAMS initiates the download of the full-resolution images from various sources to its backend. Finally, the new annotated images are merged with the existing data to form a new version of the training set. 
	
	\subsection{Training the Model}
	
		With the growing success of deep learning for visual recognition \cite{krizhevsky2012imagenet,lecun2015deep,goodfellow2016deep,he2016deep}, we use a deep convolution network as the model to recognize food images.   
	
		\subsubsection{Transferable Features for Image Recognition}
		
		In recent years, it was observed that using networks pre-trained on the ImageNet dataset and transferring those to other datasets provided a signficant boost to performance than training new models from scratch. This was due to the ability of Deep Convolution Networks to learn general features applicable to several computer vision tasks \cite{donahue2014decaf,yosinski2014transferable}. Following this success, we use pre-trained ImageNet models and fine tuned them on our food dataset. During the course of this project, we have tried several models, and updated them as new state-of-the-art models got invented. During the earlier iterations on FoodAI we tried older models such as AlexNet \cite{krizhevsky2012imagenet}, VGG \cite{simonyan2015very}, GoogLeNet \cite{szegedy2015going}. In this paper, we focus our attention to more recent models and report their performance: ResNet \cite{he2016deep}, ResNeXt \cite{xie2017aggregated} and SENet \cite{hu2018squeeze}. We also considered DenseNet \cite{huang2017densely}, but found its performance comparable to ResNet models. During training, we followed the standard approaches for data augmentation such as rotation, random crop, random contrast, etc. 

		\subsubsection{Class Imbalance in the Dataset}
		
		A specific problem we face in the FoodAI dataset is the extremely imbalanced data, i.e., there is huge variance in the number of instances per class (see histogram of instance distribution in Figure \ref{fig:dataDistribution}). This imbalance would induce a bias in the model favoring a better performance to those classes with more data, even if those food items are relatively easy to classify. To address this issue, we modify the traditional cross-entropy loss to \textit{focal loss} \cite{lin2018focal}, during the training phase. This loss will dynamically vary the scale of the loss of the instances such that the focus is more on the difficult examples during training. Specifically, instead of using cross-entropy we modify the loss as:
		$$ FL(p_t) = -\alpha_t(1-p_t)^\gamma \log (p_t)$$
		
		Here $\alpha$ is a factor that balances the weightage of samples from different classes, and $\gamma > 0$ is the focusing parameter, which regulates the importance of the sample based on its ease of classification. If the sample is easy to classify, its importance gets reduced.

	\subsection{System Architecture and Deployment}
	
		After having trained the model at the backend, we deployed the FoodAI model for production. 

		\begin{figure*}[h]
			\centering
			\includegraphics[width=0.8\textwidth]{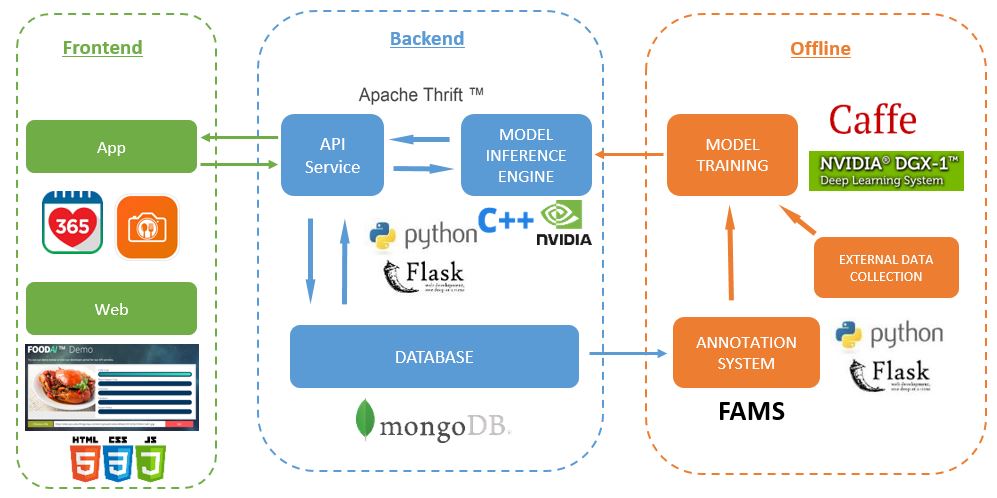}
			\caption{System Architecture for FoodAI. The end-to-end framework of deployment of FoodAI as an API service.}
			\label{fig:system}
		\end{figure*}

		\subsubsection{The System Architecture}
		
		FoodAI has been deployed as a RESTful web service accessible via HTTP/HTTPS protocol. At the front end, the client is platform and language independent. It can be a mobile app, web application, or desktop application subject to clients' business requirement. At the backend, the Apache Nginx web server is responsible for receiving user requests, redirecting to the UWSGI web application server. The UWSGI calls Caffe \cite{jia2014caffe} inference engine to get the food prediction scores and returns it to the Nginx web server, which returns the response to the user.  FoodAI web service is written in Python using Flask framework. FoodAI web service mainly provides two interfaces, classify and feedback. The classify interface receives HTTP/HTTPS request, including either the URL of image or image data, saves data and image to database and returns the classification results. The feedback interface receives the user feedback about the classification result and saves to database. In FoodAI, we use MongoDB as database. The classification interface supports both GET and POST methods. The feedback supports GET method only. To perform food image classification, the UWSGI sends image data to the Caffe inference engine. The Caffe inference engine is written in C++ and hosted on a web server. The web server loads the FoodAI model at startup time and makes inference in the GPU mode. To facilitate the cross-language communication between API service and inference engine, we use the Apache Thrift framework. The Apache Thrift framework is a scalable cross-language service solution that works efficiently and seamlessly among a various range of languages, such as Java, Python, C++ etc. An overview of the entire deployed architecture is shown in Figure \ref{fig:system}.

		\subsubsection{User Experience for API Service}
		
		To use FoodAI web service, users are required to register their interests on FoodAI website (\url{www.foodai.org}). After the requests are approved, an API key is assigned to them. The API key is required to validate user's identity for using FoodAI service. Users can integrate FoodAI web service in their application or a system independent from the platforms. The API documentation can be found at FoodAI website. The response in Json format is returned to the user. The Json object contains the attributes as follows:	
		\begin{itemize}
			\item food result: a list of top 10 visual food name sorted by the classification score.
			\item food results by category: a list of top 10 super category sorted by the classification score
			\item non-food: an indicator to show if the image is not a food
			\item qid: query id of the request
			\item status code: an indicator of the response status
			\item status message: a descriptive message based on status code
			\item time cost: the time spent on inference 
		\end{itemize}

		\subsubsection{User Experience on Healthy 365 App}
		
		The Health Promotion Board (HPB) of Singapore is one of the major partners of FoodAI project and they have integrated FoodAI web service on their mobile app called Healthy 365 on both IOS and Android platforms. HPB is a government organization committed to promoting healthy living in Singapore and Healthy 365 enables users to track their daily caloric intake and consumption. When the users want to update their diet journal, they take a photo of their meal to be identified by the FoodAI system. After the necessary computation, FoodAI returns a list of top visual food categories sorted by the classification score. The user can then choose which visual food category best describes their meal and determine further variations (e.g., coffee with or without sugar), based on the visual food chosen. Of course, if the correct item is not in the top predicted results, users have the freedom to manually enter their own choice of food. This feedback by the user is recorded for the purpose of monitoring the performance of FoodAI in the real world and to help improve the performance of the model. 

\section{Experiments and Case Studies}

	\subsection{Evaluation of Model during Development}
	
	Here we present the results of FoodAI in the training phase, where we evaluate the performance on test data of the original dataset. 
		
		\begin{table*}[htbp]
			\centering
			\caption{Performance of various models on FoodAI-756 dataset. The models were pre-trained on ImageNet and fine tuned on our dataset. Combination of SENet and ResNeXt gave a top-1 accuracy of 80.86\%. We also look at the inference (testing speed) and the model size, to help with practical decision making of trading off speed and performance. Best performance is in bold.}
			\begin{tabular}{lcccc}
				\hline
				\textbf{Network} & \multicolumn{1}{l}{\textbf{Top-1 Accuracy}} & \multicolumn{1}{l}{\textbf{Top-5 Accuracy}} & \multicolumn{1}{l}{\textbf{Testing Speed (\#Images/second)}} & \textbf{Model Size} \\
				\hline
				\textbf{ResNet-50 \cite{he2016deep}} & 0.7870 & 0.9427 & 80    & 96 MB \\
				\textbf{ResNet-101 \cite{he2016deep}} & 0.7645 & 0.9366 & 32    & 168 MB \\
				\textbf{ResNeXt-50 \cite{xie2017aggregated}}& {0.7898} & {0.9473} & 112   & 94 MB \\
				\textbf{SENet ResNeXt-50 \cite{hu2018squeeze}+\cite{xie2017aggregated}} & \textbf{0.8086} & \textbf{0.9561} & 122   & 103 MB \\
				\hline
			\end{tabular}%
			\label{tab:basic_results}%
		\end{table*}%

	\begin{table*}[htbp]
	\centering
	\caption{Performance enhancement by using Focal Loss to address issues arising from imbalanced dataset. Using Focal loss, we managed to improve our top-1 accuracy from the best fine-tuned models at 80.86\% to 83.2\%.  Best performance is in bold.}
	\begin{tabular}{lcc}
		\hline
		& \multicolumn{1}{l}{\textbf{Top-1 Accuracy}} & \multicolumn{1}{l}{\textbf{Top-5 Accuracy}} \\
		\hline
		\textbf{ResNet50 without Focal Loss} & 0.787 & 0.943 \\
		\textbf{ResNet-50 + Focal Loss} & 0.802 & 0.95 \\
		\textbf{SENet + ResNeXt50 + Focal loss} & 0.823 & 0.955 \\
		\textbf{SENet + ResNeXt101 + Focal loss} & \textbf{0.832} & \textbf{0.957} \\
		\hline
	\end{tabular}%
	\label{tab:focal_loss}%
	\end{table*}%
		
	\subsubsection{Performance of Fine-Tuning Pre-Trained ImageNet Models}
		
		While several models have been explored over the course of FoodAI development, in this paper, we present the performance of ResNet-50, ResNet-101 (50-layer and 101-layer ResNet) \cite{he2016deep}, ResNeXT-50 (50 layers) \cite{xie2017aggregated}, and SENet trained with ResNeXt-50 \cite{hu2018squeeze,xie2017aggregated}. We present the results of the basic models in Table \ref{tab:basic_results}. Among the models, the best top-1 accuracy of 80.86\% and top-5 accuracy of 95.61\% was obtained by a combination of SENet with ResNeXt-50. ResNet-101 did not do very well (possibly due to convergence challenges). We also look at the inference speed of the models and see that we can make predictions at the rate of 80-120 images per second for the 50-layer models, or 1 image in 0.01 seconds. This is fairly fast and the end-to-end inference result returned to the user thus depends on the round-trip latency of transferring the image to our server and getting the result back. The models occupy close to 100MB for the 50 layer models. Since this model is going to be stored only on the server, the model does not cause a memory constraint.


	\subsubsection{Performance After Incorporating Focal Loss}
	
	We present the results of training the model with focal loss \cite{lin2018focal} in Table \ref{tab:focal_loss}. Here, due to the usage of focal loss, we have improved the convergence of the model to a better optimum. We obtain a top-1 accuracy of 83.2\%, achieved with a combination of SENet\cite{hu2018squeeze} and ResNeXt-101 \cite{xie2017aggregated}, outperforming the previous best of 80.86\%. This demonstrates the ability of focal loss to improve the performance on imbalanced datasets like the FoodAI-756 dataset by dynamically changing the scale of the loss during training and giving less importance to easy examples.

	\subsubsection{Insights from Performance on the Test Dataset}

\begin{figure}[h]
		\centering
		\includegraphics[width=0.45\textwidth]{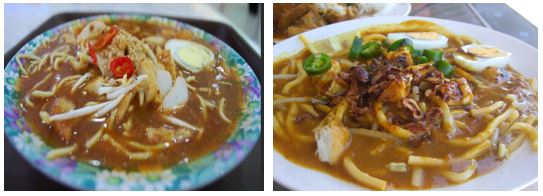}
		\caption{Example of items misclassified in development environment: \textit{Mee Kuah}(left) and \textit{Mee Rebus} (right). Best viewed in color.}
		\label{fig:confused}
		\vspace{-0.5cm}
	\end{figure}

	Next, we look closer at some of the results so as to obtain new insights. Specifically, we focused on the most misclassified instances. Our objective was to understand why these specific classes would get misclassified. Was it a possible short-coming in our training strategy? Was it that the data was just too difficult or noisy? Was it that many of these classes looked visually very similar? Or possibly a combination of any of these factors. We show some of these highly misclassified results in Table \ref{tab:confused_label}. In many of the cases, from the visual food name, we can see that some of the items and their predictions have very similar ingredients, which leads to confusion. In particular the first row in the table shows the same dish with soup predicted as being without soup (dry). One possible explanation of this is that our dataset has more instances of the dry version of the dish than the soupy version. Another interesting case is the last row: Mee Kuah, and Mee Rebus. See Figure \ref{fig:confused} for images of both classes, where we can see that they look very similar. At the time of dataset collection, we assumed that these should have been different categories. Upon further research, we found that these categories are often considered as the same item. This suggests merging them into a single visual food. 

	\begin{table*}[h]
		\centering
		\caption{Difficult Cases in the FoodAI-756 test dataset (in the development environment). These are some of the most incorrectly classified items in the test dataset. Delving deeper can give us actionable insights on how to improve the model.}
		\begin{tabular}{lcl}
			\hline
			\textbf{Visual Food} & \textbf{Recall} & \textbf{Most common incorrect prediction} \\
			\hline
			mushroom\_and\_minced\_pork\_noodles\_soup & 0.2   & dry\_minced\_pork\_and\_mushroom\_noodles \\
			vegetable\_u\_mian & 0.24  & dry\_ban\_mian \\
			stewed\_taupok & 0.28  & bak\_kut\_teh \\
			tauhu\_goreng & 0.3   & fried\_tau\_kwa \\
			pork\_chop\_western\_set & 0.32  & chicken\_chop \\
			tikka & 0.4   & chicken\_curry \\
			beef\_ball\_soup & 0.42  & beef\_ball\_kway\_teow\_soup \\
			dao\_xiao\_mian & 0.42  & been\_noodles\_soup \\
			instant\_coffee & 0.42  & kopi\_o \\
			mee\_kuah & 0.42  & mee\_rebus \\
			\hline
		\end{tabular}%
		\label{tab:confused_label}%
	\end{table*}%

	\begin{figure*}[h]
		\centering
		\includegraphics[width=\textwidth]{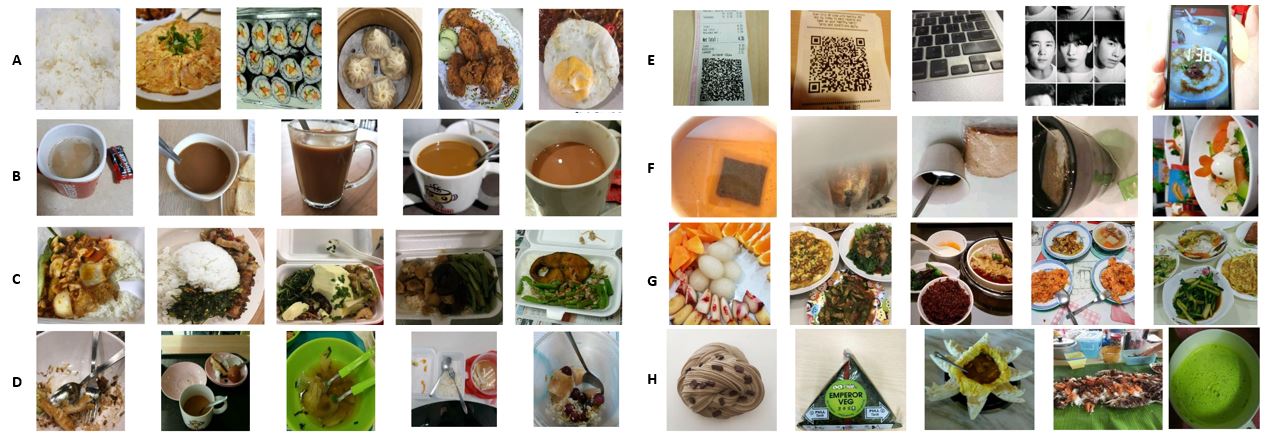}
		\caption{Examples of different types of queries sent by users, and the different types of challenges we face.\\
			A. These are some of the easy queries sent by the users, where we get good quality images, and easy to recognize classes. \\ 
			B. Large inter-class similarity. First 3 images are \textit{instant\_coffee} and the next two are \textit{teh-c/teh-o}, and they look visually similar.\\
			C. Large intra-class diversity. These are all examples of \textit{economy\_rice}, where the images look very different from one another.\\
			D. Incomplete food. Often the users will send images of food already consumed, making it difficult to detect visual features.\\ 
			E. Non Food. Being a relatively new technology, curious users will play with FoodAI and submit several non-food queries. \\
			F. Poorly taken photos. Users will submit queries where the photos taken suffer from illumination, rotation, occlusion, etc.\\
			G. Multiple Food Items. Several queries will have multiple food items present, while FoodAI is trained to detect a single class.\\
			H. Unknown Foods. We receive queries of food items that are not available in our list, making it impossible to recognize them. 
		}
		\label{fig:user_query_examples}
	\end{figure*}

	\begin{figure*}[h]
		\centering
		\includegraphics[height=4.9cm,width=\textwidth]{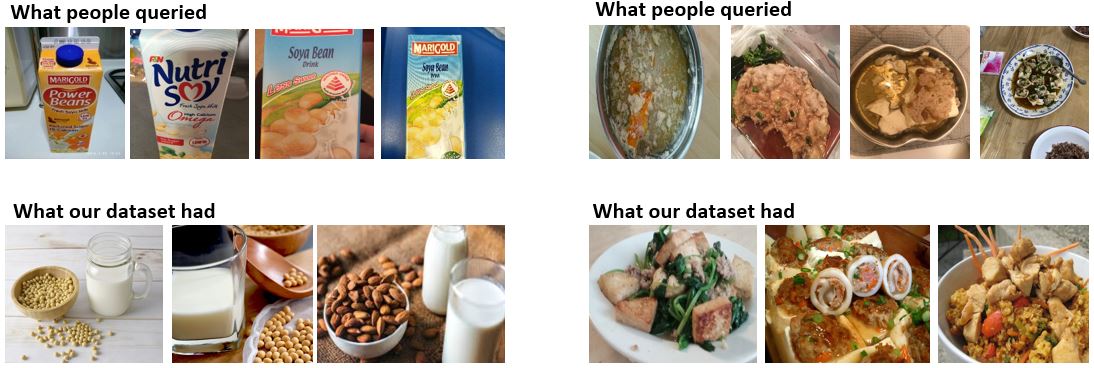}
		\caption{Case Study 1: User query data, where we had a poor performance (on both top-1 and top-5 accuracies). First column is the class \textit{soya\_milk}. While our model was trained to recognize the actual soya milk, several users queried cartons of soya milk, where "food-image features" were not prominent. Second column is the class \textit{steamed\_stir\_fried\_tofu\_with\_minced\_pork}. Here, the images in our training set were much clearer and cleaner, while the user data was not of sufficient quality for our model to extract useful visual features. Best viewed in color}
		\label{fig:case1}
		\includegraphics[height=4.9cm,width=\textwidth]{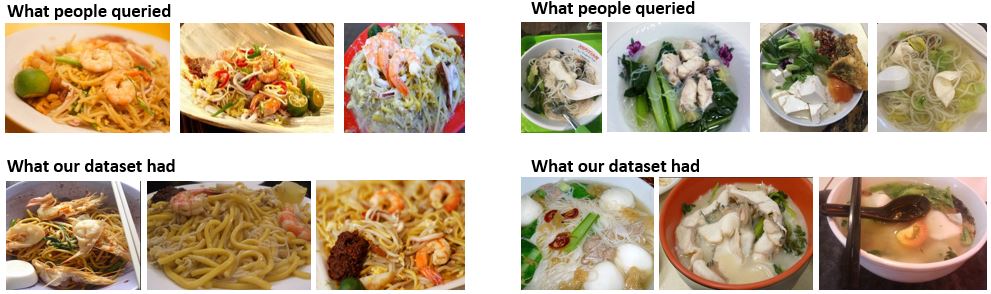}
		\caption{Case Study 2: User query data, where out top-1 accuracy was poor, but top-5 accuracy was high. These are cases where the foods have very similar features. The first column is the class \textit{dry\_prawn\_noodles}. Most of the top-1 predictions were \textit{hokkien\_mee}, which has very similar ingredients. The second column is the class \textit{fish\_beehoon\_soup}, whose top-1 prediction was \textit{beehoon\_soup\_with\_mixed\_ingredients\_eg\_seafood}. Both dishes have very similar appearances. Best viewed in color}
		\label{fig:case2}
	\end{figure*}

	\subsection{Evaluation of Model in Production}	
	
	Our deployed system receives about several API calls a day. As expected we have 3 peaks in the usage during the day, one in the morning at around 7 AM, one at lunch time, between 12 noon to 2 PM and one at dinner time from 6 PM to 8 PM. Here we present results of our analysis of the query data by the users. 
	
		\subsubsection{Performance of the Model}
		
		How do we measure the model performance in the real world? One approach could be to manually annotate data that has been queried, and compare this against the model predictions. This approach is extremely expensive for two reasons: (i) It requires substantial manual effort, which would be time-consuming; and (ii) This labeling process requires an expert who is familiar with all 756 visual food categories (for a variety of cuisines) and finding/hiring such experts for labeling is not an easy task. Note that annotating these images is significantly harder than collecting them. This is because during data collection, we just query a keyword and retrieve several images immediately which requires a cursory look to confirm. In the case of user queries, the annotator has to look at one image and assign it one of the 756 categories (some of whom are visually very similar) .

		Another approach to measure model performance is based on the feedback given by the users. Despite not receiving feedback for all queries, this is a useful indicator. Based on the feedback, we score an accuracy of about 50\% in top-1 accuracy, and about 80\% in top-5 accuracy. There could be several factors contributing to a worse performance in the real world than on our test dataset. We hypothesize the following possible (not exhaustive) reasons:
		\begin{itemize}
			\item \textbf{Domain Shift}. Possibly, the distribution of query data and our training data is different. Our data is relatively cleaner, and has higher quality photos in comparison to the real world photos taken by the users. This domain shift \cite{ganin2015unsupervised,ganin2016domain,hoffman2018cycada} may cause a degradation of performance of the model.
			\item \textbf{Poor Quality of Data Query}. Closely related to domain shift is the poor quality of image queries. Many users may submit queries which are of poor quality (e.g. upside down images, food is mostly consumed, etc.). These images would result in a poor performance of the model.
			\item \textbf{Different Imbalanced Distribution}. As noted before, the FoodAI-756 dataset is highly imbalanced, creating a bias in the model learnt. This bias may affect model performance in the real world, where the distribution of instances per class queried may be different from that in the train dataset. 
			\item \textbf{Poor Feedback Quality}. It is possible that users may not have the knowledge of food item and may gave an arbitrary feedback. They may also be intending to "play" around with the technology and intentionally giving a wrong feedback.
		\end{itemize}
		
		Next, we will explore some of these factors through case studies. 
		
		\subsubsection{Analysis of User Queries}
		We first look at some of the queries sent by users, and identify some of the key properties. We show several examples in Figure \ref{fig:user_query_examples}, where we have categorized them into 8 categories based on the associated challenges. While several queries are easy to classify (A), there are many challenges including inter-class similarity (B), intra-class diversity (C), incomplete food (D), non-food (E), poorly taken photos (F), multiple food items (G), and unknown foods (H). Detailed descriptions are in the caption. 
		
		\vspace{-0.4cm}
		
		\subsubsection{Query Images vs Our Dataset}

		Here we present a couple of case studies based on FoodAI behavior we wanted to investigate. 
		
		\textit{Case Study 1}. We wanted to explore some cases where we had a very poor performance on the user query data (while the performance on test data in the development stage was reasonable). We considered two cases: \textit{soya\_milk} and \textit{steamed\_stir\_fried\_ tofu\_with\\\_minced\_pork}. We have visualized some the samples queried by the users and the ones in our FoodAI-756 dataset in Figure \ref{fig:case1}. In the case of \textit{soya\_milk}, it was clear that the query data was completely different from our data, as users queried cartons, while our model was trained on actual milk. Thus, the result was not surprising, and it tells us that we should potentially account for any text in packaged goods to try to improve predictions. In the case of \textit{steamed\_stir\_fried\_ tofu\_with\_minced\_pork}, it would appear that there is a domain shift in the query set and our data, which might possibly be overcome by domain adaptation techniques \cite{ganin2016domain,hoffman2018cycada}.
		
		\textit{Case Study 2}. We also wanted to explore the scenario where the top-1 accuracy was poor but the top-5 accuracy was very high. Such scenarios would give us insight into the most confusing food items according to the users (where our performance is consistently good, but we struggle to give the best prediction), and also highlight cases where the users maybe potentially giving us incorrect feedback. In Figure \ref{fig:case2}, we show two examples: (i) \textit{dry\_prawn\_noodles} was almost always top-1 predicted as \textit{hokkien\_mee}. Both have similar ingredients, and it is possible the model was biased due to a different number of instances for each class in our training set, as opposed to the user query data; (ii) We have a similar explanation for \textit{fish\_beehoon\_soup}, whose top-1 prediction was mostly \textit{beehoon\_soup\_with\_mixed\_ingredients\_eg\_seafood}. Note that it is very hard for us to validate the class imbalance hypothesis as that requires us to manually label a large amount of the query data. 
	
\vspace{-0.2cm}
\section{Conclusions and Future Directions}

	We have developed FoodAI, a deep learning based food image recognition system for smart food logging. FoodAI helps reduce the burdens of manually logging an online food journal by facilitating photo-based food journals. The system has been trained to identify 756 different types of foods, specifically covering a variety of cuisines commonly consumed in Singapore. We have conducted several experiments to train a powerful model relying on state-of-the-art visual recognition methods, and further improved the performance by incorporating focal loss. We have presented analysis of how we updated the dataset regularly, and how we obtained actionable insights based on the model performance during development. The technology has been deployed, and we have several organizations and universities using this service. One of our major partners Health Promotion Board, Singapore has integrated FoodAI into the Healthy 365 App. We get over several API calls a day. We have also conducted extensive analysis and case studies to obtain insights into the performance of the model in the real world. 
	
	We are also pursuing several research and development directions. One of the major challenges is how to update the model to incorporate new classes of food as they become popular. Retraining the model can be very expensive and time consuming. To alleviate this, we are exploring Lifelong Learning solutions \cite{parisi2018continual}. A related idea is if the data for new classes is very limited, how do we extend the model to recognize this class? We are looking into incorporating few-shot learning techniques to do this \cite{lake2015human}. Since there are several food items, it will not be possible for us to maintain a fully exhaustive list. Another way to incorporate calorie consumption is to estimate calories directly from the image. A related task that we are exploring is cross-modal retrieval between food images and cooking recipes \cite{salvador2017learning}, where we want to retrieve the recipe for a given image (and it is easier to estimate nutrition and calorie consumption from recipes) . We are also looking at incentivization strategies for healthier consumption and food recommendation \cite{ge2015health}. We are making efforts to expand FoodAI research into a viable solution for aiding smart consumption and a healthy lifestyle. 

\section{Acknowledgments}
This research is supported by the National Research Foundation, Prime Minister's Office, Singapore under its International Research Centres in Singapore Funding Initiative. We would also like to acknowledge collaboration with Health Promotion Board, Singapore.

\bibliographystyle{ACM-Reference-Format}
\bibliography{bibliography}

\end{document}